%% file: elsarticle-template.tex
\documentclass[review]{elsarticle}

\usepackage{hyperref}

\usepackage{xcolor}
\usepackage{amssymb}
\usepackage{multirow}
\usepackage{epsfig}
\usepackage{colortbl}
\usepackage{algorithm2e}
\usepackage{xspace}
\usepackage[small]{caption}
\usepackage{graphicx}
\usepackage{amsmath}
\usepackage{booktabs}

\newcommand{\eg}{\textit{e.g., }}
\newcommand{\ie}{\textit{i.e., }}
\newcommand{\etal}{\textit{et al.}}

\journal{Journal of \LaTeX\ Templates}

\begin{document}

\begin{frontmatter}

\title{Medi-Care AI: Predicting Medications From Billing Codes via Robust Recurrent Neural Networks}


\author[address1]{Deyin Liu}
\ead{iedyzzu@outlook.com}
\author[address2,address3]{Lin Wu \corref{correspondingauthor}}
\cortext[correspondingauthor]{Corresponding author}
\ead{xiaoxian.wu9188@gmail.com}
\author[address4]{Xue Li}
\ead{lixue@neusoft.edu.cn}




%
%

\address[address1]{Zhengzhou University, China}
\address[address2]{Key Laboratory of Knowledge Engineering with Big Data (Hefei University of Technology), Ministry of Education}
\address[address3]{School of Computer Science and Information Engineering, Hefei University of Technology, Hefei 230000, China}
\address[address4]{Dalian Neusoft University of Information, China}

\begin{abstract}
In this paper, we present an effective deep prediction framework based on robust recurrent neural networks (RNNs) to predict the likely therapeutic classes of medications a patient is taking, given a sequence of diagnostic billing codes in their record. Accurately capturing the list of medications currently taken by a given patient is extremely challenging due to undefined errors and omissions. We present a general robust framework that explicitly models the possible contamination through overtime decay mechanism on the input billing codes and noise injection into the recurrent hidden states, respectively. By doing this, billing codes are reformulated into its temporal patterns with decay rates on each medical variable, and the hidden states of RNNs are regularised by random noises which serve as dropout to improved RNNs robustness towards data variability in terms of missing values and multiple errors. The proposed method is extensively evaluated on real health care data to demonstrate its effectiveness in suggesting medication orders from contaminated values.
\end{abstract}


\end{frontmatter}


\section{Introduction}\label{sec:intro}

There has been growing interest in exploiting the large amounts of data existed in electronic medical records for both clinical events and secondary research. While leveraging large historical data in electronic health records (EHR) holds great promise, its potential is weakened by multiple errors and omissions in those records. Some studies show that over 50\% of electronic medication lists contain omissions \cite{Caglar2011-emergency}, and even 25\% of all medications taken by patents are not recorded. To ensure the corrections of medication lists, great efforts have been dedicated to improve the communications between patients and providers \cite{Keogh}, however, manually maintaining these lists would be extremely human-labor intensive. Thus, it demands a generic yet robust predictive model that is able to suggest medication consultation to the patients next visit in the context of medication documentation contaminations.

Recently, Recurrent Neural Networks (RNNs), such as Long Short-Term Memory (LSTM) \cite{lstm1997}, and Gated Recurrent Unit (GRU) \cite{GRU2014} have been explored for modeling diseases and patient diagnosis in health care modality \cite{LSTM-ICU,Doctor-AI,MiME,SRL-RNN}. For instance, a temporal model based on RNNs, namely Doctor AI, is developed to predict future physician diagnosis and medication orders. This intelligent system demonstrates that historical EHR data can be leveraged to forecast the patient status at the next visit and present medication to a physician would like to refer at the moment. However, little efforts are put into systematically modelling the EHR with missing values \cite{GRU-D} since it is difficult to capture the missing patterns in medical billing codes. Simple solutions such as omitting the missing data and to perform analysis only on the observed data, or filling in the missing values through smoothing/interpolation \cite{Effect-missing}, spectral analysis \cite{Gappy-time-series,Yang-TNNLS18,TIP17Yang,IJCAI-16,NN18Yang}, and multiple imputations \cite{Multi-imputation} offer plausible ways to the missing values in data series. However, these solutions often result in suboptimal analysis and poor predictions because the imputations are disparate from the prediction models and missing patterns are not properly described \cite{Missing-EHR}.

A recent finding demonstrates that missing values in time series data are usually \textit{informative missing}, that is, the missing values and patterns are related to the target labels in supervised learning tasks. For example, Che \etal \cite{GRU-D} show that the missing rates in time series health care data are usually highly correlated with the labels of interests such as mortality and ICD-9 diagnoses. Hence, it demands an appropriate strategy to describe the decaying on diagnostic measurements over time. Moreover, the diagnostic billing codes are characterized of more than missing values in patient records, whereas in most cases they are combined with multiple errors and omissions. Thus, we use the terminology \textit{noise} to generally refer to all potential incorrectness of medication lists.

\subsection{Our Approach}

Inspired by the noise-based regularizer of RNNs, a.k.a dropout \cite{Wager2013-Dropout,Dropout}, we impose a multiplicative noise into the hidden states to ensure the robustness of recurrence and also preserve the underlying RNN in the context of noise injection. Hence, in this paper we develop a robust RNN model, an effective new way to deal with incomplete billing codes in medical domain whilst being capable of predicting the future medication orders given the missing codes in sequence. The key idea is to not only model the input codes by explicitly encoding the missing patterns over time, but also inject random noise into the transition function of recurrence. Intuitively, the explicit noise injection into the hidden states of RNNs can serve as regulariser to drop the observation difference that will be potentially added into the hidden states. Thus, the RNNs are trained to fit its parameters to maximize the corresponding marginal likelihood of observations in the context of high variability. The proposed model is experimentally evaluated on real EHR datasets to demonstrate its effectiveness in identifying missing actual information in relation to therapeutic classes.

\subsection{Contributions}

The contributions of this paper can be summarized as follows.
\begin{itemize}
\item We present a robust RNN based medication prediction framework to effectively cope with sequential billing codes that are contaminated by missing values and multiple errors.
\item The proposed approach is designed to predict the complete set of medications a patient is actively taking at a given moment from a sequence of diagnostic billing coeds in the context of non-trivial billing record noise. This is, to our best knowledge, the first effort to \textit{explicitly} model both the medication care data and delving the RNNs into the medical domain.
\item Insightful analysis to our approach are provided in this paper. Extensive experiments on health care datasets are conducted to demonstrate the superiority of our method over state-of-the-art by achieving the performance gain on AUC by 13\% and 7\% on the Physio-net challenge dataset \cite{Physio-Net} and MIMIC-III \cite{Mimic-iii}, respectively.
\end{itemize}

The rest of this paper is organized as follows. Section \ref{sec:related} reviews some related works. We detail the proposed predictive model in Section \ref{sec:approach} with some background described in Section \ref{sec:background} in advance. Section \ref{sec:exp} reports extensive experiments over the real-valued medical datasets, and the paper is concluded in Section \ref{sec:con}.

\input{related.tex}
\input{approach.tex}

\input{experiment.tex}

\section{Conclusions and Future Work}\label{sec:con}

In this paper, we present an effective approach to medicare system, which is a RNN-based deep learning model that can learn robust patient representation from a large amount of longitudinal patient billing code records and predict future medication lists. We demonstrate the effectiveness of our method which achieved improved recall accuracy values in the real medical practice with observed missing values or incorrect records. In the future work, we would strive to improve the performance the recurrent networks by including additional input data, such as laboratory test results, demographics, and perhaps vital signs related to rare diseases. One interesting direction is to figure out a pathway to convert the medication data into reliably-ordered sequences, so as to fully exploit the strength of recurrent networks for medication prediction.


\bibliography{mybibfile}

\end{document}

%% file: related.tex
\section{Related Work}\label{sec:related}

\subsection{Modeling medical event sequences}

Common approaches to modeling medical event sequences include continuous-time Markov chain based models \cite{Bayesian-markov} and their extension using Baysian networks \cite{Multiplicative-forests} as well as intensity function methodologies such as Hawkes processes \cite{Disease-hawkes}. It is known that continuous-time Markov chain methods are computationally expensive because modeling multi-labelled point processes would expand rapidly their state-space. On the other hand, Hawkes processes with intensity functions depend linearly with respect to the past observations, while they are limited in capturing temporal dynamics. Moreover, there is no study on these models to deal with missing values or incorrect data. In this paper, we address these challenges by designing a robust recurrent neural network which has shown to be effective in learning complex yet potentially missing data in sequential patterns regarding health-care systems.

\subsection{Deep learning models for EHR}

It has witnessed some attempts to apply neural network models a.k.a deep learning methods to study EHR since deep learning models are capable of learning complex data patterns. The earlier work is the use of an LSTM model that produced reasonable accuracy (micro-AUC 0.86) in a 128-dim multi-label prediction of diagnoses from regularly sampled, continuously real-valued physiologic variables in an Intensive Care Unit (ICU) setting \cite{LSTM-ICU}. One successful framework is Doctor AI \cite{Doctor-AI} which is a predictive temporal model using RNNs to predict the diagnosis and medication codes for a subsequent visit of patients. They used a GRU model in a multi-label context to predict the medications, billing codes, and time of the next patient visit from a sequence of that same information for previous visits. It can achieve an improvement over a single-hidden-layer MLP (reach a recall$@30$ of 70.5 by a 20 margin). This is a successful showcase of using the strength of recurrence,\ie to predict the next element in a sequence. However, aforementioned deep learning paradigms are not able to effectively cope with EHR with errors and omissions.

Prior efforts have been dedicated into modeling missing data in sequences with RNNs in clinical time series \cite{Lipton2016-missing-rnn,GRU-D,Missing-EHR}. A very recent work yet contemporary with our work, namely GRU-Decay \cite{GRU-D}, used a GRU model with imputation on missing data by a decay term to predict the mortality/ICD-9 diagnosis categories from medication orders and billing codes. Our method contrasts with GRU-Decay \cite{GRU-D} in the way of managing the RNN to tackle the missing values. Instead of using the same decay mechanism on both input sequence and the hidden state as the GRU-Decay performed, we propose to dealing with the raw inputs and hidden states in different strategies wherein the input billing codes are multiplied by decay rate on each variable (the same as GRU-Decay \cite{GRU-D}), and the hidden states are injected into noises in the multiplicative form. 

%% file: approach.tex
\section{Background}\label{sec:background}

\subsection{Medical billing codes}

In our experiments, codes are from the International Classification of Disease, Ninth Revision (ICD-9). The ICD-9 hierarchy consists of 21 chapters roughly corresponding to a single organ system or pathologic class. Leaf-level codes in the tree represent single diseases or disease subtypes. For each time a patient has billable contact with the health-care system through which the time stamped billing codes are associated with the patient record, indicating the medical conditions that are related to the reasoning for the visit. However, these billing codes are more often unreliable or incomplete, and thus making the electronic medical records unable to track the set of medications that the patient is actively taking. The code range and descriptions are shown in Table \ref{tab:ICD-9}.

\begin{table}
  \centering
  \begin{center}
  \begin{tabular}{|c c |}
  \hline
  \textbf{Code range} & \textbf{Description} \\
  \hline
  001-139 & Infectious and parasitic diseases  \\
  140-239 & Neoplasms \\
  240-279 & Endocrine, nutritional and metabolic diseases, immunity disorders\\
  280-289 & Blood diseases and blood-forming organs\\
  290-319 & Mental disorders\\
  320-359 & Nervous system diseases\\
  360-389 & Sense system diseases\\
  390-459 & Circulatory system diseases\\
  460-519 & Respiratory system diseases\\
  520-579 & Digestive system diseases\\
  580-629 & Genitourinary system diseases\\
  630-679 & Complications of pregnancy, childbirth, and the puerperium \\
  680-709 & Skin and subcutaneous tissue\\
  710-739 & Musculoskeletal system and connective tissue\\
  740-759 & Congenital anomalies \\
  760-779 & Conditions originating in the perinatal period\\
  780-799 & Symptoms, signs and ill-defined conditions\\
  800-999 & Injury and poisoning\\
 \hline
\end{tabular}
\end{center}
  \caption{The top level classes for ICD-9 chapters.}\label{tab:ICD-9}
\end{table}

\subsection{Recurrent neural networks}

An recurrent neural network (RNN) considers a sequence of observations, $\mathbf{X}_{1:T}=(\mathbf{x}_1,\ldots, \mathbf{x}_T)$, and to handle the sequential time-series the RNN introduces the hidden state $\mathbf{h}_t$ at time step $t$, as a parametric function $f_W(\mathbf{h}_{t-1}, \mathbf{x}_{t-1})$ of the previous state $\mathbf{h}_{t-1}$ and the previous observation $\mathbf{x}_{t-1}$. The parameter $W$ is shared across all steps which would greatly reduce the total number of parameters we need to learn. The function $f_W$ is the transition function of the RNN, which defines a  recurrence relation for the hidden states and renders $\mathbf{h}_t$ a function of all the past observations $\mathbf{x}_{1:t-1}$.

The particular form of $f_W$ determines the variants of RNN including Long-Short Term Memory (LSTM) \cite{lstm1997} and Gated Recurrent Units (GRU) \cite{GRU2014}. In this paper, we will study GRU which has shown very similar performance to LSTM but employs a simper architecture. First, we would reiterate the mathematical formulation of GRU as follows
\begin{equation}\label{eq:GRU}
\begin{split}
& \mathbf{z}_t = \sigma(\mathbf{W}_z \mathbf{x}_t + \mathbf{U}_z \mathbf{h}_{t-1}), \mathbf{r}_t = \sigma(\mathbf{W}_r \mathbf{x}_t + \mathbf{U}_r \mathbf{h}_{t-1}),\\
& \hat{\mathbf{h}}_t = tanh(\mathbf{W}_h \mathbf{x}_t + \mathbf{U}_h (\mathbf{r}_t \odot \mathbf{h}_{t-1})),\\
& \mathbf{h}_t = (1 - \mathbf{z}_t)\odot  \mathbf{h}_{t-1} + \mathbf{z}_t \odot  \hat{\mathbf{h}}_t,
\end{split}
\end{equation}
where $\odot$ is an element-wise multiplication. $\mathbf{z}_t$ is an update gate that determines the degree to which the unit updates its activation. $\mathbf{r}_t$ is a reset gate, and $\sigma$ is the sigmoid function. The candidate activation $\hat{\mathbf{h}}_t$ is computed similarly to that of traditional recurrent unit. When $\mathbf{r}_t$ is close to zero, the reset gate make the unit act as reading the first symbol of an input sequence and forgets the previously computed state.

\section{Robust Recurrent Neural Networks for Medication Predictions}\label{sec:approach}

In this section, we develop a new framework for clinical medication predictions in the context of missing information and multiple errors. We first formulate the prediction problem setting, and then detail the architecture with explicit noise injection into the recurrent hidden states. Finally, we present the training procedure on the proposed model.

\subsection{Problem setting}

For each patient, the temporal observations are represented by multivariate time series with $D$ variables of length $T$ as $\mathbf{X}_{1:T}\in \mathbb{R}^{T\times D}$, where $\mathbf{x}_t \in \mathbb{R}^D$ denotes the $t$-th observations, namely measurements of all variables and $x_t^d$ denotes the $d$-th variable of $\mathbf{x}_t$. In the medication records, the variables correspond to multiple medication codes, such as the codes 493 (asthma) and 428 (heat failure) from ICD-9. For each time stamp, we may extract high-level codes for prediction purpose and denote it by $\mathbf{y}_t$. Generic Product Identifier (GPI) medication codes are extracted from the medication orders. This is because the input ICD-9 codes are represented sequentially while the medications are represented as a list that changes over time. Also, many of the GPI medication codes are very granular, for example, the pulmonary tuberculosis (ICD-9 code 011) can be divided into 70 subcategories (011.01, 011.01,...,011.95, 011.96).

In this paper, we are interested in learning an effective vector representation for each patient from his billing codes over time with multiple missing values at each time stamp $t=1,\ldots,T$, and predicting diagnosis and medication categories in the next visit $\mathbf{y}_{T+1}$. We investigate the use of RNN to learn such billing code representations, treating the hidden layers as the representation for the patient status and use them for the prediction tasks. To account for the situation of missing/incorrect values in EHR, we propose robust RNN architecture, which effectively models the missing patterns from time series onwards through the temporal decay mechanism \cite{Zhou2007,Vodovotz2013,GRU-D} and injects noises into the hidden states of RNN at each time step.

\subsection{Robust RNNs with noise injection}

To effectively learn representations from missing or incorrect values in billing codes, we propose to incorporate different strategies in regards to the input billing codes and the hidden states, respectively. For the missing values in billing codes of EHR, we employ the decay mechanism which has been designed for modeling the influence of missing values in health care domain \cite{Vodovotz2013}. This is based on the property that the values of missing variables tend to be close to some default value if its last measurement is observed a long time ago. This property should be considered as critical for disease diagnosis and treatment. Also, the influence of the input dimensions will fade away over time if some dimension is found missing for a while. On the other hand, the hidden states of RNNs should be injected with random noises which is more advantageous by preventing the dimensions of hidden states from co-adapting and it can force the individual units to capture useful features \cite{Noisein}.

Specifically, we inject a decay rate into each variable of the billing code series. In this way, the decay rate differs from variable to variable and indicative to unknown possible missing patterns. To this end, the vector of a decay rate is formulated as
\begin{equation}\label{eq:decay-rate}
  \gamma_t = \exp\{-\max( \mathbf{0}, \mathbf{W}_{\gamma} \mathbf{\delta}_t + \mathbf{b}_{\gamma} )\},
\end{equation}
where $\mathbf{W}_{\gamma}$ and $\mathbf{b}_{\gamma}$ are trainable parameters jointly with the LSTM. $\exp\{\cdot\}$ is the exponential negative rectifier to keep each decay rate monotonically decreasing ranged between 0 and 1. $\delta_t^d$ is the time interval for each variable $d$ since its last observation, which can be defined as
\begin{equation}\label{eq:time-interval}
\delta_t^d = \left\{
\begin{array}{l}
     s_t - s_{t-1} + \delta_{t-1}^d, t>1 \\
     0, t=1
\end{array}\right.
\end{equation}
In Eq.\eqref{eq:time-interval}, $s_t$ denotes the time stamp when the $t$-th observation is obtained and we assume that the first observation is made at time $t=0$ ($s_1=0$). Hence, for a missing variable code, we adopt the decay vector $\gamma_t$ to decay it overtime but towards an empirical mean instead of using its last observation. And the decaying measurement billing code vector can be formulated by applying the decay scheme into:
\begin{equation}\label{eq:decay-code}
  x_t^d \leftarrow \gamma_{x_t}^d x_{t'}^d + (1- \gamma_{x_t}^d) \hat{x}^d,
\end{equation}
where $x_{t'}^d$ is the last observation of the $d$-th variable ($t'<t$) and $\hat{x}^d$ is the empirical mean of the $d$-th variable. We remark that when the input billing code is decaying, the parameter $\mathbf{W}_{\gamma_x}$ should be constrained to be diagonal so as to ensure the decay rates of variables are not affecting each other.

To augment the RNN's capability of coping with multiple errors in sequential EHR billing codes, we explicitly redefine the hidden states by injecting noises. This strategy is able to effectively fit the parameters of RNN by maximizing the likelihood of data observations because the next predicted output from RNN is determined as $p(\mathbf{x}_t|\mathbf{x}_{1:t-1})=p(\mathbf{x}_t|\mathbf{h}_t)$ \footnote{The likelihood $p(\mathbf{x}_t|\mathbf{h}_t)$ can be in the form of the exponential family.}. Thus, we define the GRU with noise as follows
\begin{equation}\label{eq:noise-GRU}
\begin{split}
& \epsilon_{1:T} \thicksim \{0, (1-\delta)\}^d;\\
& \mathbf{h}_t = f_W(\mathbf{x}_{t-1},\mathbf{h}_{t-1}, \epsilon_t) = (1 - \mathbf{z}_t)\odot  \mathbf{h}_{t-1} \odot \epsilon_t + \mathbf{z}_t \odot  \hat{\mathbf{h}}_t \odot \epsilon_t.
\end{split}
\end{equation}
In Eq.\eqref{eq:noise-GRU}, the noise component $\epsilon_{1:T}$ is an independent drawn from a scaled Bernoulli (1-$\delta$) random variable. In this paper, it is used to create the dropout noise via the element-wise product ($\cdot$) of each time hidden state $\mathbf{h}_{t_1}$. In other words, dropout noise corresponds to setting $\mathbf{h}$ to 0 with probability $\delta$, and to $\mathbf{h}/(1-\delta)$ else. Intuitively, this multiplicative form of noise injection can induce the robustness of RNN to how future data may be different from the observations. Also, this can be regarded as a regularization on RNN to normalize the hidden states, which is similar to noise-based regularizer for neural networks, namely dropout \cite{Wager2013-Dropout,Dropout}. This explicit regularization is equivalent to fitting the RNN loss to maximize the likelihood of the data observations, while being with a penalty function of its parameters. This type of regularization that involves noise variables can help the RNNs learn long-term dependencies in sequential data even in the context of high variability because dropout-based regularization can only drop differences that are added to network's hidden state at each time-step. And thus this dropout scheme allows up to use per-step sampling while still being able to capture the long-term dependencies \cite{Recurrent-dropout}.

\subsection{The architecture of the prediction model}

\begin{figure}[t]
\begin{tabular}{cc}
\hspace{-8mm}\includegraphics[width=2.5in,height=1.5in]{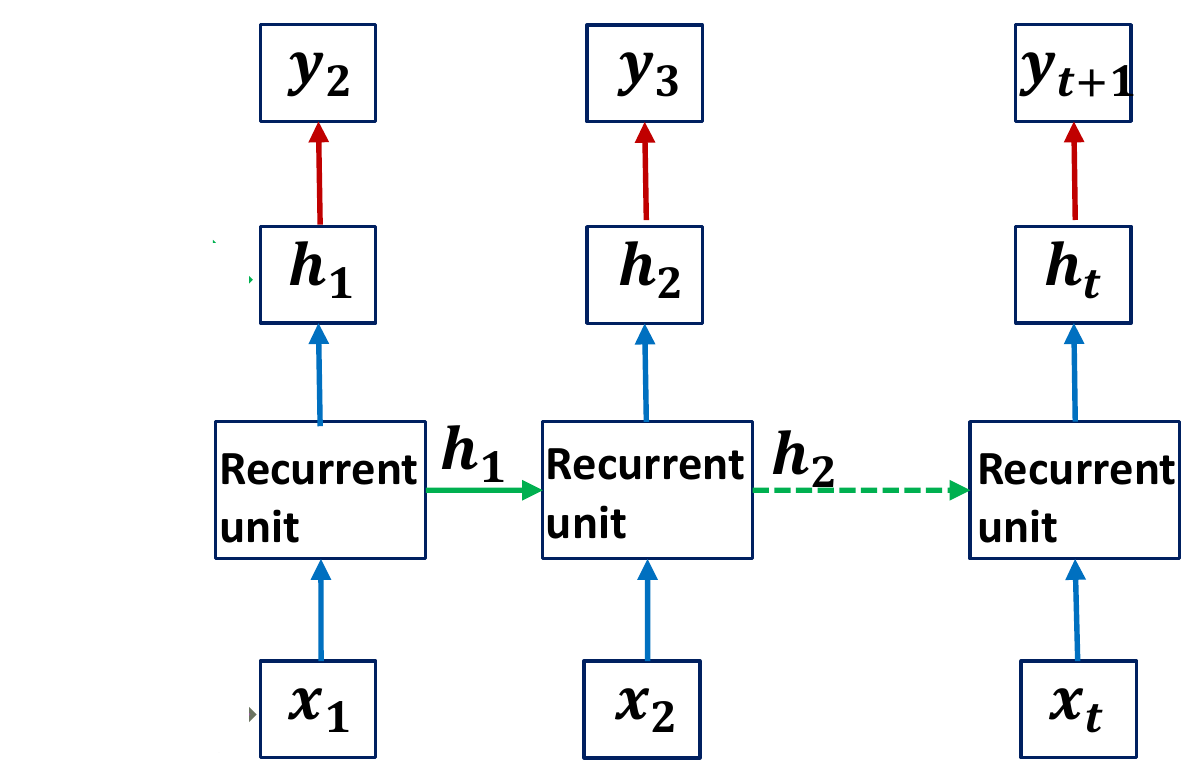}&
\includegraphics[width=2.5in,height=1.5in]{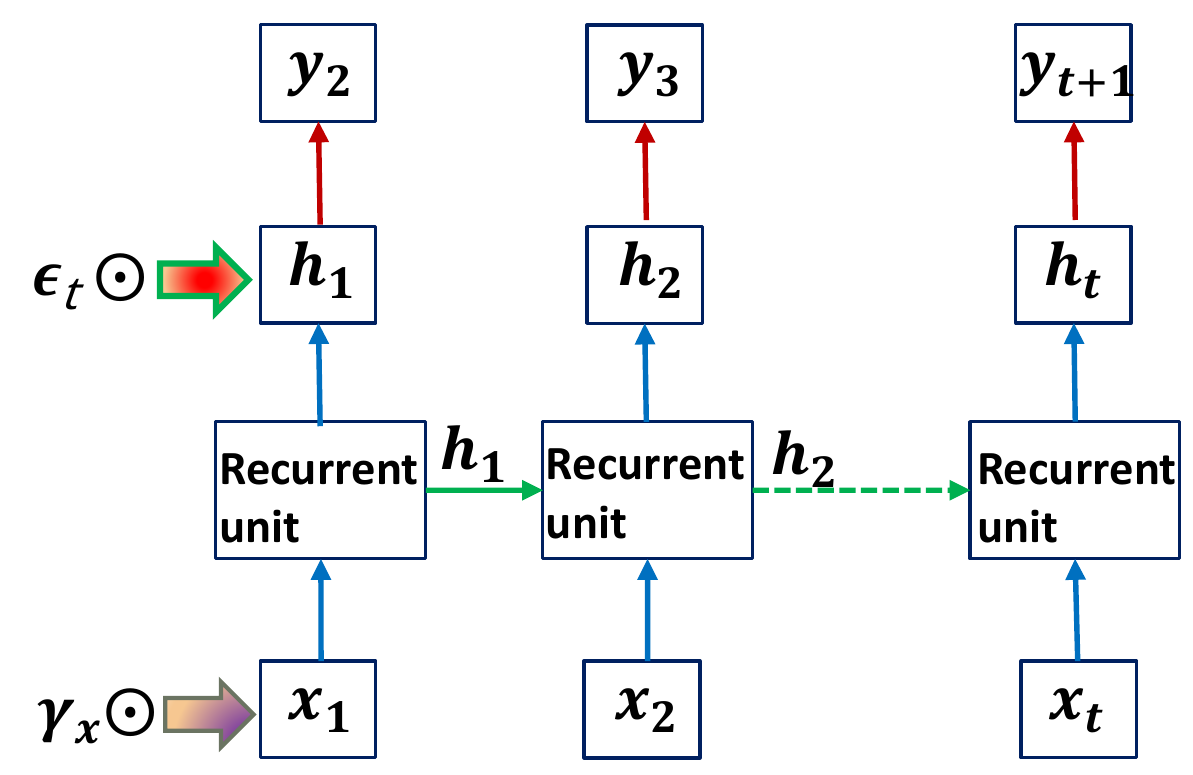}\\
(a) & (b)
\end{tabular}
\caption{The overview of our framework with robust RNN to solve the problem of forecasting the medication codes assigned to a patient for his next visit. (a) A conventional RNN model. (b) The proposed model. The input sequential data in regards to a patient ($\mathbf{X}=\{\mathbf{x}_1,\ldots,\mathbf{x}_t\}$) are embedded with decay mechanism ($\gamma_{x}$) to model the potential missing pattern, and the stacked recurrent layers with multiplicative noise regularization ($\epsilon_t$) learn the status of the patient at each time stamp. Given the learnt status ($\mathbf{h}_t$), the framework is to generate the codes observed in the next time stamp.}\label{fig:architecture}
\end{figure}

As shown in Fig.\ref{fig:architecture}, the proposed robust neural network architecture receives input at each time stamp $t$ corresponding to patient visits in sequences. The billing codes $\mathbf{x}_t$ are in the form of multi-label categories. The input sequential billing codes are modelled with the decay of missing values, and then fed into the stacked multiple layers of GRU to project the inputs into lower dimensional space, and also learn the status of the patients at each time stamp as real-valued vectors. For predicting the diagnosis codes and the medication codes at each time stamp $t$, a softmax layer is stacked on top of the GRU, using the hidden state $\mathbf{h}_t$ as the input, that is, $\mathbf{y}_{t+1}=softmax(\mathbf{W}_{code}^T \mathbf{h}_t + \mathbf{b}_{code})$. Thus, the objective of our model is to learn the weights $\mathbf{W}_{[z,r,h,code, \gamma]}, \mathbf{U}_{[z,r,h]}, \mathbf{b}_{[z,r,h,code, \gamma]}$. In particular, the values of all $\mathbf{W}$ and $\mathbf{U}$ are initialized to orthogonal matrices using singular value decomposition of matrices from the normal distribution \cite{Saxe2013}. All values of $\mathbf{b}$ are initialized to be zeros. Therefore, for each patient we employ the cross entropy as the loss function for the code prediction, which is defined as
\begin{equation}\label{eq:loss-function}
  \mathcal{L}(\mathbf{W}, \mathbf{U}, \mathbf{b})=\sum_{t=1}^{n-1} \left( \mathbf{\ddot{y}}_{t+1} \log (\mathbf{y}_{t+1}) + (1-\mathbf{\ddot{y}}_{t+1}) \log (1-\mathbf{y}_{t+1}) \right),
\end{equation}
where $\mathbf{\ddot{y}}$ is the ground truth medication category.

\begin{algorithm}[t]
\KwData{Input billing codes in sequence $\mathbf{x}_{1:T}$, initial hidden state $\mathbf{h}_0$, noise distribution $\varphi(\cdot:1,\sigma)$.}
\KwResult{Set of learned parameters of GRU: $\mathbf{W}_{[z,r,h,code,\gamma]}, \mathbf{U}_{[z,r,h]}, \mathbf{b}_{[z,r,h,code,\gamma]}$. }
Initialize the set of parameters  \;
\While{stopping criterion not met}{
    \For{$t$ from 1 to $T$}{
    Sample noise from $\epsilon_t \thicksim \varphi(\epsilon_t:1,\sigma)$ \;
    Compute the decayed inputs $x_t^d \leftarrow \gamma_{x_t}^d x_{t'}^d + (1- \gamma_{x_t}^d) \hat{x}^d$ \;
    Compute state $\mathbf{h}_t = (1 - \mathbf{z}_t)\odot  \mathbf{h}_{t-1} \odot \epsilon_t + \mathbf{z}_t \odot \hat{\mathbf{h}}_t \odot \epsilon_t$ \;
    }
    Compute loss as in Eq. \eqref{eq:loss-function} \;
    Update the network parameters \;
  }
\caption{The proposed robust RNN framework for medication prediction from billing codes.}
\end{algorithm}

%% file: experiment.tex
\section{Experiments}\label{sec:exp}

In this section, we demonstrate the performance of our model on two real-world health-care datasets, and compare it to several strong machine learning and deep learning competitors in the classification tasks.

\subsection{Data preparation and experimental setting}

We conduct experiments on two health-care datasets: Physio-net challenge dataset \cite{Physio-Net} and MIMIC-III \cite{Mimic-iii}.

\begin{itemize}
  \item Physio-net challenge 2012 dataset (Physio-Net): This PhysioNet Challenge dataset \cite{Physio-Net} is a publicly available collection of multivariate clinical time series from 8,000 intensive care unit (ICU) records. Each record is a multivariate time series of roughly 48 hours and contains 33 variables such as albumin, heart-rate, glucose etc. We use the training subset A in our experiments since ground truth outcomes are only publicly available on this subset. We conduct the prediction of 4 tasks on this dataset: in-hospital mortality, length-of-stay less than 3 days, had a cardia condition or not, and whether the patient was recovering from surgery. This can be treated as a multi-task classification problem.
  \item MIMIC-III: This is a publicly available dataset collect at Beth Israel Deaconess Medical Center from 2001 to 2012 \cite{Mimic-iii}. It contains over 58,000 hospital admission records, and we extract 99 time series features from 19,714 admission records for 4 modalities which are very useful for monitoring ICU patients \cite{GRU-D}. These modalities include input-events (fluids to patients, \eg insulin), output-events (fluids out of the patient, \eg urine), lab-events (lab test results, \eg pH values), and prescription-events (active drugs prescribed by doctors, \eg aspirin). We use the fist 48 hours data after admission from each time series, and conduct the predictive ICD-9 code task: predict ICD-9 diagnostic categories (\eg respiratory system diagnosis) for each admission, which can be treated as a multi-label problem.
\end{itemize}

For the training on all the models, we use 85\% of the patients as the training set, and 15\% as the testing set. All the RNN models are trained with 50 epoches \ie 50 iterations over the entire training data, and then evaluate the performance against the testing set. To avoid over-fitting, we apply the dropout between the GRU layer and the final prediction layer, and also between the multiple stacked GRU layers. The dropout rate is 0.3 and the norm-2 regularization is applied into the weight matrix of $\mathbf{W}_{code}$.  The dimensionality of the hidden states $\mathbf{h}$ of the GRU is set to be 2048 to ensure the expressive power. We train the models using truncated back-propagation through time with average stochastic gradient descent \cite{Average-SGD}. To avoid the problem of exploding gradients, we clip the gradients to a maximum norm of 0.25.

\subsection{Evaluation metrics}

For the evaluation on the task in a multi-label context, the performance of all methods is evaluated against two metrics: the micro-averaged area under the ROC curve (AUC) and the top-k recall. The measure of AUC treats each instance with equal weight, regardless of the nature of the positive labels for that instance \cite{Bajor2017}, which would not give a score advantage to instances with very prevalent or very rare labels. The micro-averaged AUC considers each of the multiple label predictions as either true or false, and then computes the binary AUC if they all belong to the same 2-class problem, Thus, the micro-average AUC $\mathcal{A}_{\mu}$ can be defined as

\begin{equation}\label{eq:AUC}\small
\mathcal{A}_{\mu}=\frac{|(\mathbf{x},\mathbf{x'},l,l'): f(\mathbf{x},l)\geqslant f(\mathbf{x'},l'), (\mathbf{x},l) \in \mathcal{S}, (\mathbf{x'},l') \in \mathcal{\bar{S}}|}{|\mathcal{S}||\mathcal{S'}|}
\end{equation}
where $\mathcal{S}=\{\mathbf{x},l\}: l\in Y$ is the set of [instance, label] pairs with a positive label, and $Y=\{y_d:y_d=1,\ldots, D\}$ is the set of positive labels for the input $\mathbf{x}$.

The top-$k$ recall mimics the behavior of doctors examining differential diagnosis which suggest the doctor is listing most probable diagnoses and treat the patients accordingly to identity the patients status. The top-k recall is defined as
\begin{equation}\label{eq:recall}\small
  \text{top}-k \, \text{recall} = \frac{\# \text{TP in the top k predictions}}{\#\text{TP}},
\end{equation}
where $\#\text{TP}$ denotes the number of true positives. Thus, a machine with high top-k recall translates to a doctor with effective diagnostic skills. In this end, it turns out to make top-k recall a suitable measure for the performance of prediction models on medications.

\subsection{Baselines}

We consider baselines in two categories: (1) RNN based methods: Doctor-AI \cite{Doctor-AI}, GRU-Decay \cite{GRU-D}, LSTM-ICU \cite{LSTM-ICU}; MiME \cite{MiME}; SRL-RNN \cite{SRL-RNN}; (2) Non-RNN based methods: Logistic Regression (LR), Support Vector Machine (SVM), and Random Forest (RF).

\begin{itemize}
 \item Doctor-AI \cite{Doctor-AI}: Doctor AI is a temporal model using RNN to assess the history of patients to make multi-label predictions on physician diagnosis and the next medication order list.
\item GRU-Decay \cite{GRU-D}: To tackle the missing values in EHR data, GRU-Decay is based on Gated Recurrent Units and exploits the missing patterns for effective imputation and improves the prediction performance.
 \item LSTM-ICU \cite{LSTM-ICU}: It is a study to empirically evaluate the ability of LSTMs to recognize patterns in multivariate time series of clinical measurements. They consider multi-label classification of diagnoses by training a model to classify 128 diagnoses given frequently but irregularly sampled clinical measurements.
 \item MiME \cite{MiME} A Multilevel Medical Embedding (MiME) approach to learn the multilevel embedding of EHR data that only relys on this inherent EHR structure without the need for external labels.
 \item SRL-RNN \cite{SRL-RNN} A Supervised Reinforcement Learning with Recurrent Neural Network (SRL-RNN), which fuses them into a synergistic learning framework.
\item Logistic Regression (LR): Logistic regression is a common method to predict the codes in the next visit $\mathbf{x}_t$ using the past $\mathbf{x}_{t-1}$. Following \cite{Doctor-AI}, we use the data from $L$ time lags before and aggregate the data $\mathbf{x}_{t-1}+\mathbf{x}_{t-2}+,+\mathbf{x}_{t-L}$ for some duration $L$ to create the feature for prediction on $\mathbf{x}_t$.
 \item Support Vector Machine (SVM): A multi-label SVM is trained to obtain multiple classifiers to each diagnostic code and each medication category.
\item Random Forest (RF): The random forest is not easily constructed to work on sequences, and we represented the input data as bag-of-code vector $b\in \mathbb{R}^D$. As RF cannot be operated on large-size dataset, we break down it into an ensemble of ten independent forests while each one trained on one tenth of the training data, and their averaged score is used for test prediction.
\end{itemize}

\subsection{Results and discussions}

\paragraph{Prediction performance}

\begin{table}[t]
  \centering
  \scriptsize
  \begin{tabular}{|l|l|c|c|}
    \hline
     \multicolumn{2}{|c|}{\cellcolor{gray} \textcolor{white} {Method}} & \cellcolor{gray} \textcolor{white} {Physio-Net} & \cellcolor{gray} \textcolor{white} {MIMIC-III} \\
     \hline
   \parbox[t]{1mm}{\multirow{5}{*}{\rotatebox[origin=c]{90}{RNN}}} & Ours & \color{red}$\mathbf{0.90}$ & \color{red}$\mathbf{0.78}$ \\
   \cline{2-4}
   & Doctor-AI \cite{Doctor-AI} & 0.77 & 0.71 \\
   & GRU-Decay \cite{GRU-D} & 0.84 & 0.76 \\
   & LSTM-ICU \cite{LSTM-ICU} & 0.76 & 0.70 \\
   & SRL-RNN \cite{SRL-RNN} & 0.86 & 0.74\\
    \hline
   \parbox[t]{1mm}{\multirow{6}{*}{\rotatebox[origin=c]{90}{Non-RNN}}} & Logistic Regression & 0.64 & 0.66\\
   & SVM  & 0.71 & 0.69 \\
   & Random Forest  & 0.71 & 0.73 \\
   & Logistic Regression-mean &0.66 & 0.67\\
   & SVM-mean & 0.72 & 0.71 \\
   & Random Forest-mean & 0.72& 0.73\\
    \hline
  \end{tabular}
  \caption{Comparison results of AUC on the real-valued datasets for multi-task predictions.}\label{tab:AUC}
\end{table}

In the first experiment, we evaluate all methods on Physio-Net and MIMIC-III datasets. Table \ref{tab:AUC} shows the prediction performance of al the models on the multi-task predictions on real datasets: all 4 tasks on Physio-Net and 20 ICD-9 code tasks on the MIMIC-III. The proposed method achieves the best AUC score across all tasks on both the datasets. We notice that all RNN models perform better than non-RNN methods because the deep recurrent layers help these models capture the temporal relationship that is useful in solving prediction tasks. Moreover, explicitly modelling the missing values in both the input signals and the hidden states, such as GRU-Decay and our method, can further improve the prediction results due to the capability of fitting the parameters robust to noisy time-series data.

Table \ref{tab:recall-MIMIC} compares the results of the proposed method with different algorithms in three settings: predicting only the diagnosis codes (Dx), predicting only the medication codes (Rx), and jointly predicting both Dx and Rx codes. The experimental results show that the proposed method is able to outperform the baseline algorithms by a noticeable margin. The results also confirm that RNN based approaches achieve superior performance to non-RNN methods. This is mainly because RNNs are able to learn succinct feature representations of patients by accumulating the relevant information from their history visits and the current set of codes, which outperform the hand-crafted features of Non-RNN baselines. Moreover, in the case of missing values and incorrectness in billing codes, our method achieves the best results on all measures in the merit of explicit modelling on billing code variables and robust improvement on recurrence.

\begin{table*}[hbt]
  \centering
  \small
  \begin{tabular}{|l|l|c|c|c|c|c|c|c|c|c|}
    \hline
     \multicolumn{2}{|c|}{\cellcolor{gray} \textcolor{white} {Method}} & \multicolumn{3}{|c|}{\cellcolor{gray} \textcolor{white} {Dx Recall $@k$}} & \multicolumn{3}{|c|}{\cellcolor{gray} \textcolor{white} {Rx Recall $@k$}} & \multicolumn{3}{|c|}{\cellcolor{gray} \textcolor{white} {[Dx, Rx] Recall $@k$}}\\
     \hline
     &  & $k$=10 & $k$=20 & $k$=30 & $k$=10 & $k$=20 & $k$=30 & $k$=10 & $k$=20 & $k$=30\\
     \hline
   \parbox[t]{1mm}{\multirow{6}{*}{\rotatebox[origin=c]{90}{RNN}}} & Ours & \color{red}$\mathbf{71.2}$ & \color{red}$\mathbf{77.8}$ & \color{red}$\mathbf{85.1}$ & \color{red}$\mathbf{77.2}$ & \color{red}$\mathbf{86.2}$ & \color{red}$\mathbf{92.0}$ & \color{red}$\mathbf{59.8}$ & \color{red}$\mathbf{73.5}$ & \color{red}$\mathbf{80.2}$\\
   \cline{2-11}
   & Doctor-AI \cite{Doctor-AI} & 64.3 & 74.3 & 79.6 & 68.2 & 79.7 & 85.5 & 55.0 & 66.3 & 72.5\\
   & GRU-Decay \cite{GRU-D} & 67.4 & 75.9 & 82.6 & 73.5 & 83.7 &89.0& 57.4 & 69.0 &76.1\\
   & LSTM-ICU \cite{LSTM-ICU} & 63.7 & 74.3 & 79.5 & 68.0 & 79.1 & 84.8 & 54.8 & 62.9 & 72.4\\
   & MiME \cite{MiME} & - & - & - & - & - & - & - & - & -\\
   & SRL-RNN \cite{SRL-RNN} & - & - & - & - & - & - & - & - & -\\
    \hline
   \parbox[t]{1mm}{\multirow{6}{*}{\rotatebox[origin=c]{90}{Non-RNN}}} & Logistic Regression & 43.2 & 54.0 & 60.8 & 45.8 & 60.0 & 69.0 & 36.0 & 46.3 & 52.5\\
   & SVM  &46.2 & 57.9 & 65.1 & 47.8 & 63.4 & 69.9 & 38.0 & 48.5 & 56.1\\
   & Random Forest  & 47.8 & 58.9 & 67.2 & 48.6 & 63.5 & 69.8 & 37.8 & 48.0 & 55.0\\
   & Logistic Regression-mean & 44.7 & 55.1 & 62.4 & 46.2 & 60.8 & 69.7 & 37.0 & 46.5 & 52.8\\
   & SVM-mean &46.8 & 59.6& 66.0 & 49.2 & 65.4 & 71.0 & 39.8 & 49.6 & 57.8\\
   & Random Forest-mean & 48.2 & 59.7& 67.3 & 49.0 & 65.0 & 71.1 & 39.0 & 48.2 & 55.7\\
    \hline
  \end{tabular}
  \caption{Comparison results of accuracy in forecasting future medical activities on the MIMIC-III dataset.}\label{tab:recall-MIMIC}
\end{table*}

\begin{figure}[t]
\centering
\begin{tabular}{c}
\includegraphics[height=5cm]{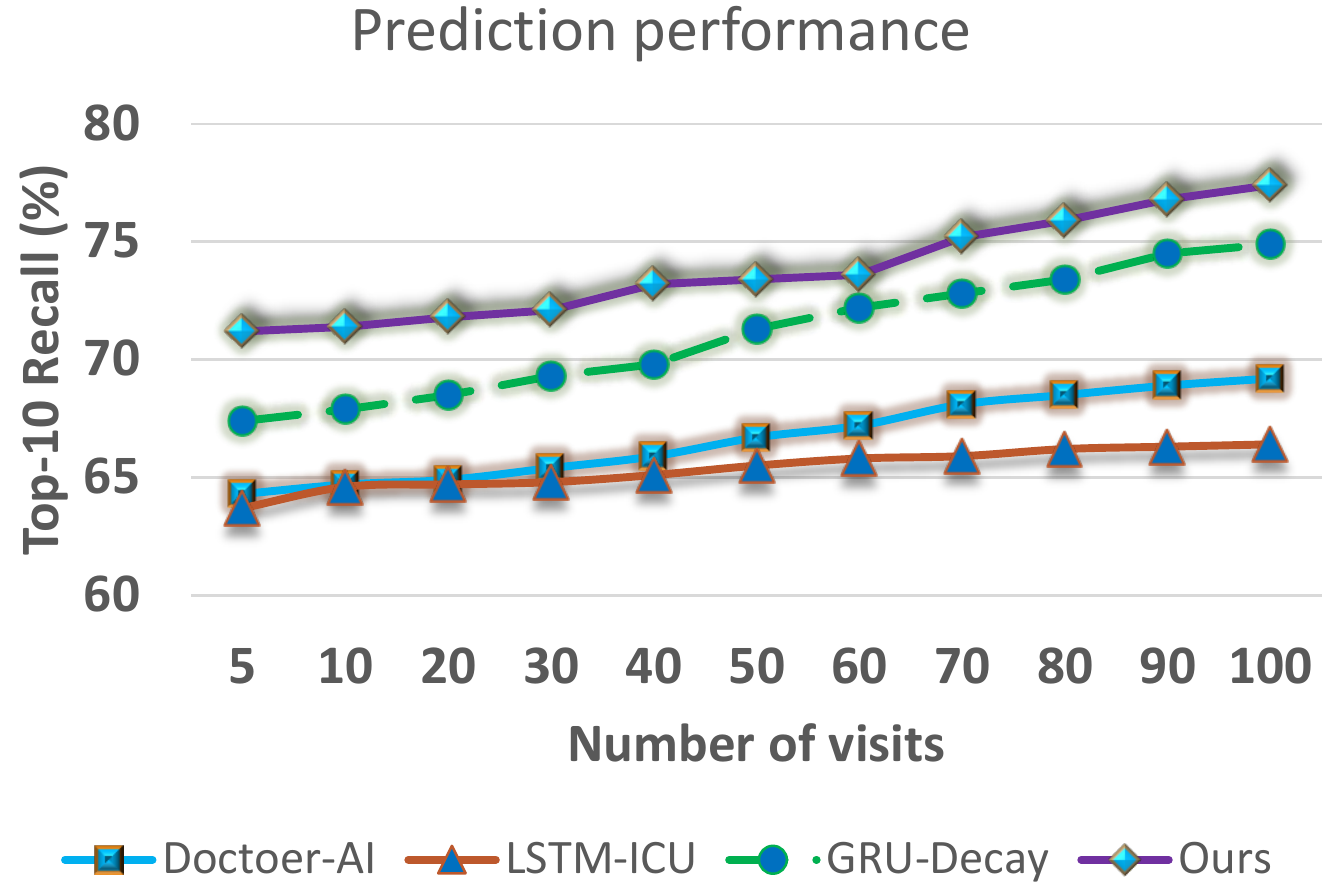}
\end{tabular}
\caption{The prediction performance with respect to the duration of patient medical history.}
\label{fig:prediction-visits}
\end{figure}

To further examine the capability of our method in a real-world medical care setting where patients may have varying lengths of their medical records, we conduct an experiment to study the affect of billing code history duration on the prediction performance. To this end, we select 5,800 patients from MIMIC-III who had more than 100 visits. We consider the RNN based deep models to predict the diagnosis codes at visit at different times and calculate the mean values of recall$@10$ across the selected patient samples. Fig.\ref{fig:prediction-visits} shows the experimental results of different RNN based models. It can be observed that all methods are increasing their performance on prediction as they see longer patient visit records, and certainly our approach achieved the best prediction performance amongst all RNN-based models. This is mainly because the recurrence is well-suited to time-series and the prediction is more faithful given longer sequence inputs. Also, it is inferred that those patients with high visit count are more likely caught in severely ill, and therefore their future is easier to predict.

\paragraph{More discussions}

As the spread $\sigma$ controls the noise level and determines the amount of regularisation into RNN, we discuss on the property of different noise distributions, \ie Gaussian and Bernoulli, and the impact on the training of RNN. The experimental results are reported in Table \ref{tab:noise-level}. It can be found that what really matters with different distributions is the variance $\sigma$ which determines the degree of regularisation into the RNN. And the RNN regularisation is not very sensitive to different types of distribution, for example, on both the health-care datasets the AUC values with Gaussian distribution are very similar to Bernoulli while for each specific distribution the spread $\sigma$ affects the performance.

\begin{table}[ht]\scriptsize
\caption{The study on different noise distributions. The micro-averaged AUC values are reported on two datasets.}
\begin{center}
\begin{tabular}{cccc}
    \hline
    Distribution & $\sigma$ & Physio-Net & MIMIC-III\\
    \hline
    \multirow{4}{*}{Gaussian}& 0.53 & 0.82 & 0.70\\
    & 0.92 & 0.86 & 0.73 \\
    & 1.10 & 0.90 & 0.78 \\
    & 1.50 & 0.87 & 0.75 \\
    \hline
    \multirow{4}{*}{Bernoulli}& 0.33 & 0.79 & 0.71\\
    & 0.41 & 0.84 & 0.72\\
    & 0.50 & 0.89 & 0.75 \\
    & 0.80 & 0.87 & 0.72 \\
    \hline
\end{tabular}
\end{center}
\label{tab:noise-level}
\end{table}

\begin{figure}[t]
\centering
\begin{tabular}{c}
\includegraphics[height=3.5cm]{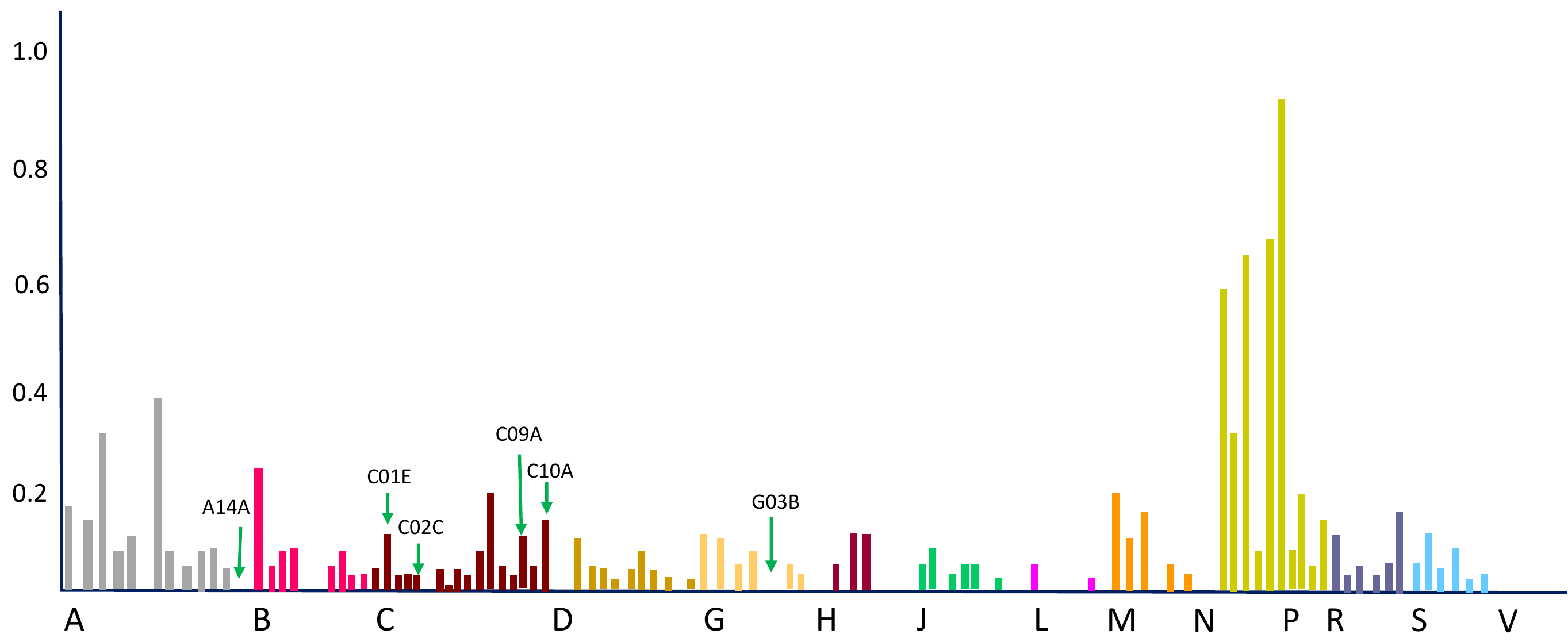}
\end{tabular}
\caption{The medication predictions for a patient with only one ICD-9 code. Each vertical bar represents the prediction for a single medication class and the height indicates the confidence of the prediction. See the texts for details.}
\label{fig:medication-prediction}
\end{figure}


To further examine the capability of our model in predicting medications in missing billings, we study a case on a patient with Parkinson's disease in which his/her record has at least five years of data consisting of only codes for Parkinson's disease whereas the data contains medications for high cholesterol, hypertension without explicit labels referring to Parkinson's disease. In fact, the medication entities listed as true labels are not suggested for paralysis agitans (Parkinson's disease), while the patient was surely taking them even though not documented into the ICD-9 sequence. As shown in Fig.\ref{fig:medication-prediction}, in the case of missing medication items, the model is still able to predict reasonable medications for a patient with Parkinson's disease, such as Dopaminergic agents and Antiepileptics, which are primary treatment for the disease. The top prediction probabilities and missing true labels on each treatment regarding a patient are reported in Table \ref{tab:Top-True}. Thus, our model is useful to identifying missing medications in the clinical scenario, such as reconciling information in a large scale from a range of electronic and human sources to establish the ground truth of medications that are taken on a particular day.


\begin{table}[t]\scriptsize
  \centering
  \begin{center}
  \begin{tabular}{|c c c|}
  \hline
\multicolumn{2}{|c|}{\cellcolor{gray} \textcolor{white} {Top predictions}} & \textcolor{red} {Prob.}\\
 \hline
 N04B & Dopaminergic agents & 98.2\% \\
 N03A & Antiepileptics & 38.4\% \\
 N02B & Other analgesics and antipyretics & 35.7\% \\
 N06A & Antidepressants & 31.2\% \\
 N02A & Opioids & 24.7\% \\
 \hline
 \multicolumn{2}{|c|}{\cellcolor{gray} \textcolor{white} {True labels}} & \textcolor{red} {Prob.}\\
 \hline
 C10A & Lipid modifying agents, plain & 17.4\%\\
 C09A & Ace inhibitors, plain & 13.2\% \\
 C01E & Other cardiac preparations & 7.8\% \\
 C02C & Antiadrenergic agents, peripherally acting & 3.7\%\\
 G03B & Androgens & 3.1\%\\
 A14A & Anabolic steroids & 2.4\%\\
 \hline
\end{tabular}
\end{center}
  \caption{A case study: Top prediction and true labels for a patient with Parkinson's disease.}\label{tab:Top-True}
\end{table}